\documentclass{article}
\PassOptionsToPackage{numbers, compress}{natbib}
\usepackage[preprint]{neurips_2021}
\usepackage{amsmath}

\usepackage[utf8]{inputenc} % allow utf-8 input
\usepackage[T1]{fontenc}    % use 8-bit T1 fonts
\usepackage{hyperref}       % hyperlinks
\usepackage{url}            % simple URL typesetting
\usepackage{booktabs}       % professional-quality tables
\usepackage{amsfonts}       % blackboard math symbols
\usepackage{nicefrac}       % compact symbols for 1/2, etc.
\usepackage{microtype}      % microtypography
\usepackage{xcolor}         % colors

\usepackage{multirow}
\usepackage{booktabs}
\usepackage{adjustbox}
\usepackage{array}

\usepackage{caption} 
\usepackage{subcaption}
\usepackage{wrapfig}
\usepackage{enumitem}

\usepackage{xspace}
\usepackage{amssymb}
\usepackage{pifont}
\usepackage{wrapfig}

\makeatletter
\DeclareRobustCommand\onedot{\futurelet\@let@token\@onedot}
\def\@onedot{\ifx\@let@token.\else.\null\fi\xspace}

\def\ie{\emph{i.e}\onedot}

\makeatother

\newcommand{\xmark}{\ding{55}}

\title{Space-time Mixing Attention for Video Transformer}

\author{%
  Adrian Bulat \\
  Samsung AI Cambridge\\
  \texttt{adrian@adrianbulat.com} \\
   \And
   Juan-Manuel Perez-Rua \\
   Samsung AI Cambridge \\
   \texttt{j.perez-rua@samsung.com} \\
   \AND
   Swathikiran Sudhakaran \\
   Samsung AI Cambridge \\
   \texttt{swathikir.s@samsung.com} \\
   \And
   Brais Martinez \\
   Samsung AI Cambridge \\
   \texttt{brais.a@samsung.com} \\
   \And
   Georgios Tzimiropoulos \\
   Samsung AI Cambridge \\
   Queen Mary University of London\\
   \texttt{g.tzimiropoulos@qmul.ac.uk} \\
}

\begin{document}

\maketitle

\begin{abstract}

This paper is on video recognition using Transformers. Very recent attempts in this area have demonstrated promising results in terms of recognition accuracy, yet they have been also shown to induce, in many cases, significant computational overheads due to the additional modelling of the temporal information. In this work, we propose a Video Transformer model the complexity of which scales linearly with the number of frames in the video sequence and hence induces \textit{no overhead} compared to an image-based Transformer model. To achieve this, our model makes two approximations to the full space-time attention used in Video Transformers: (a) It restricts time attention to a local temporal window and capitalizes on the Transformer's depth to obtain full temporal coverage of the video sequence. (b) It uses efficient space-time mixing to attend \textit{jointly} spatial and temporal locations without inducing any additional cost on top of a spatial-only attention model. We also show how to integrate 2 very lightweight mechanisms for global temporal-only attention which provide additional accuracy improvements at minimal computational cost. We demonstrate that our model produces very high recognition accuracy on the most popular video recognition datasets while at the same time being significantly more efficient than other Video Transformer models. Code will be made available.

\end{abstract}

\section{Introduction}

Video recognition -- in analogy to image recognition -- refers to the problem of recognizing events of interest in video sequences such as human activities. Following the tremendous success of Transformers in sequential data, specifically in Natural Language Processing (NLP)~\citep{vaswani2017attention,chen2018best}, Vision Transformers were very recently shown to outperform CNNs for image recognition too~\citep{yuan2021tokens,dosovitskiy2020image,touvron2020training}, signaling a paradigm shift on how visual understanding models should be constructed. In light of this, in this paper, we propose a Video Transformer model as an appealing and promising solution for improving the accuracy of video recognition models. 

A direct, natural extension of Vision Transformers to the spatio-temporal domain is to perform the self-attention \textit{jointly} across all $S$ spatial locations and $T$ temporal locations. Full space-time attention though has complexity $O(T^2S^2)$ making such a model computationally heavy and, hence, impractical even when compared with the 3D-based convolutional models. As such, our aim is to exploit the temporal information present in video streams while minimizing the computational burden within the Transformer framework for efficient video recognition. 

A baseline solution to this problem is to consider spatial-only attention followed by temporal averaging, which has complexity $O(TS^2)$. Similar attempts to reduce the cost of full space-time attention have been recently proposed in~\citep{bertasius2021space,arnab2021vivit}. These methods have demonstrated promising results in terms of video recognition accuracy, yet they have been also shown to induce, in most of the cases, significant computational overheads compared to the baseline (spatial-only) method due to the additional modelling of the temporal information. 

\begin{figure}
    \centering
    \begin{subfigure}[t]{0.23\textwidth}
        \centering
        \includegraphics[width=3.3cm]{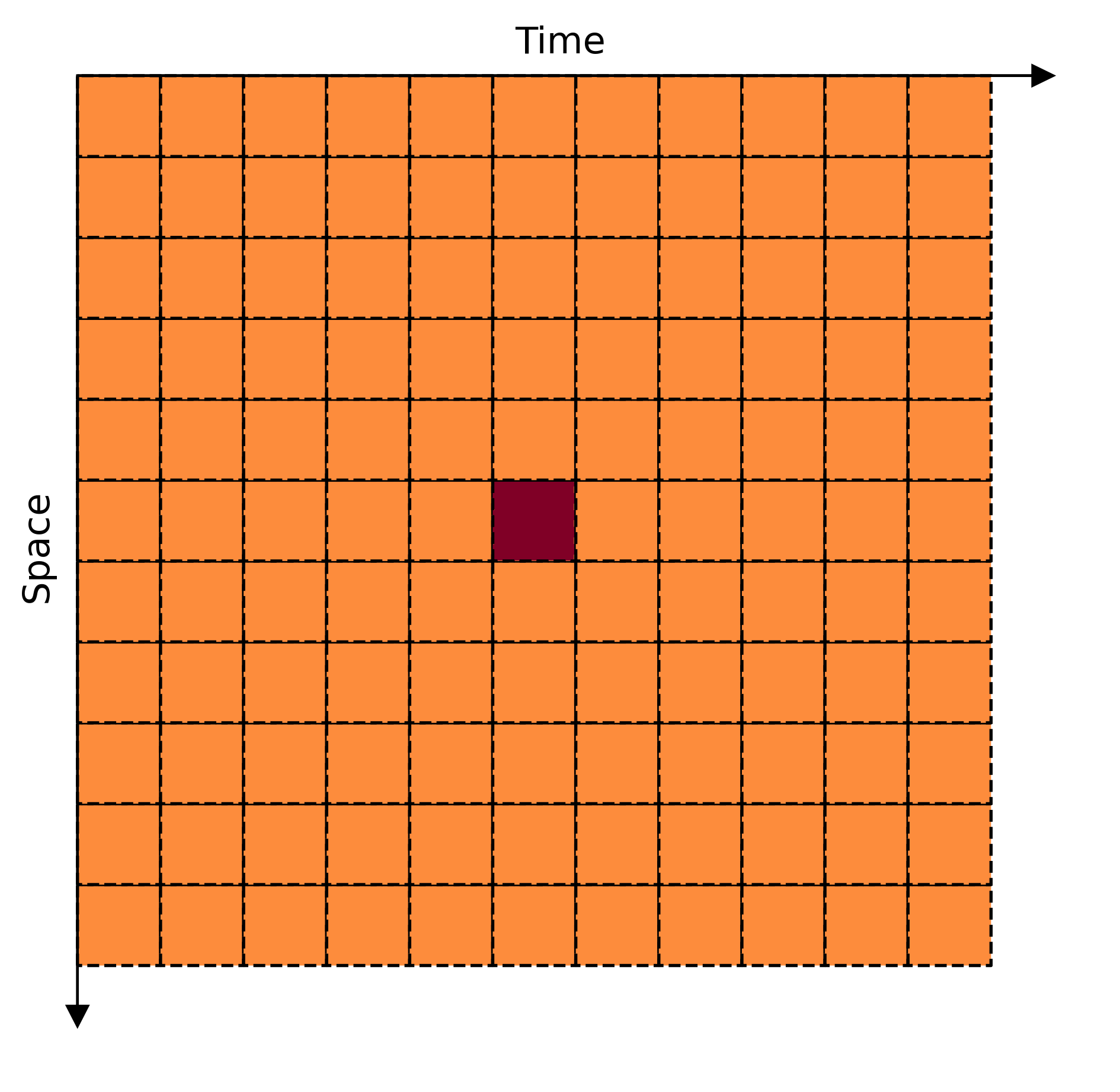}
        \caption{Full space-time attention: $O(T^2S^2)$}
    \end{subfigure}
    ~
     \begin{subfigure}[t]{0.23\textwidth}
    \centering
        \includegraphics[width=3.3cm]{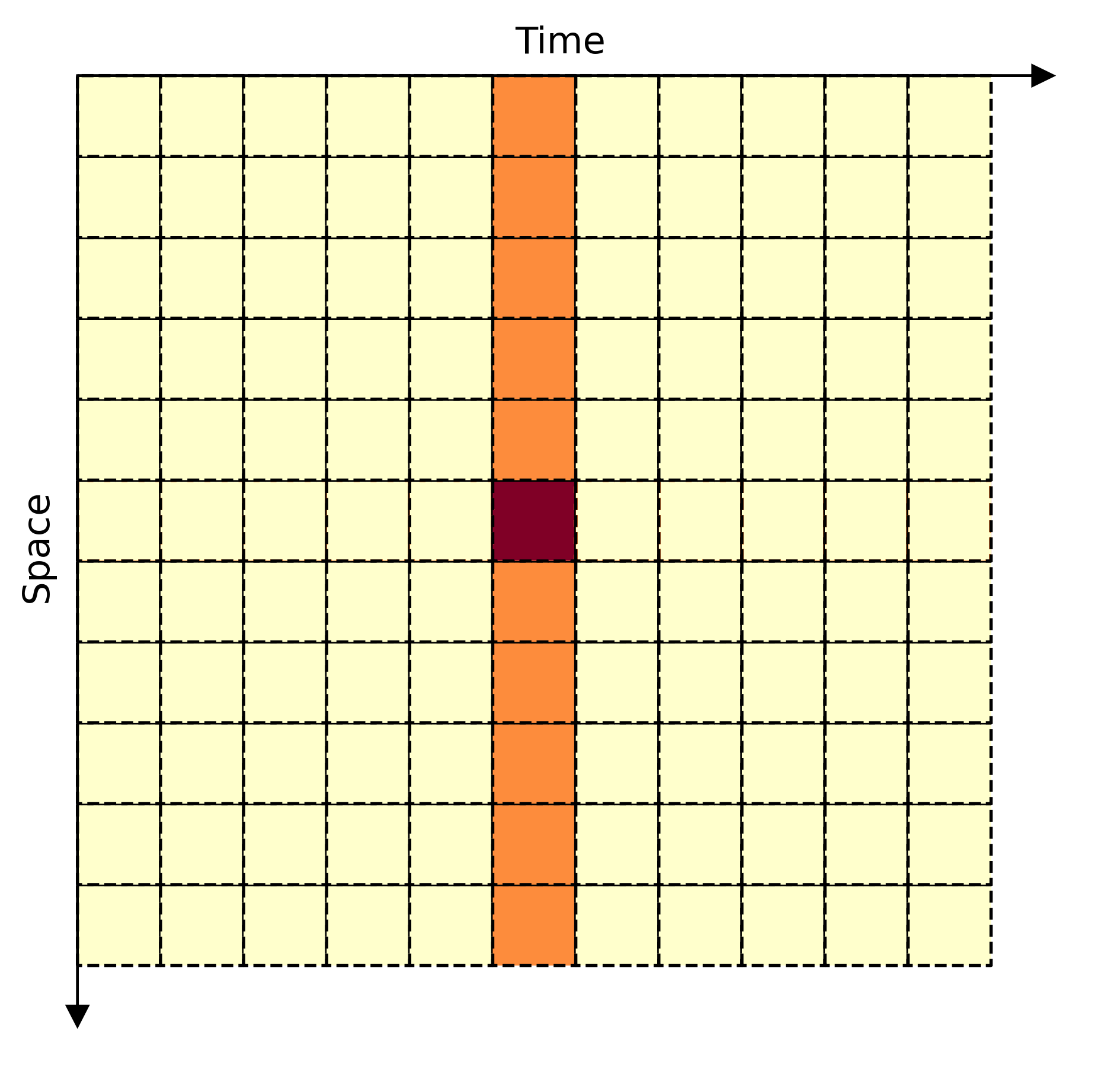}
        \caption{Spatial-only attention: $O(TS^2)$}
    \end{subfigure}
     ~
    \begin{subfigure}[t]{0.23\textwidth}
        \centering
        \includegraphics[width=3.3cm]{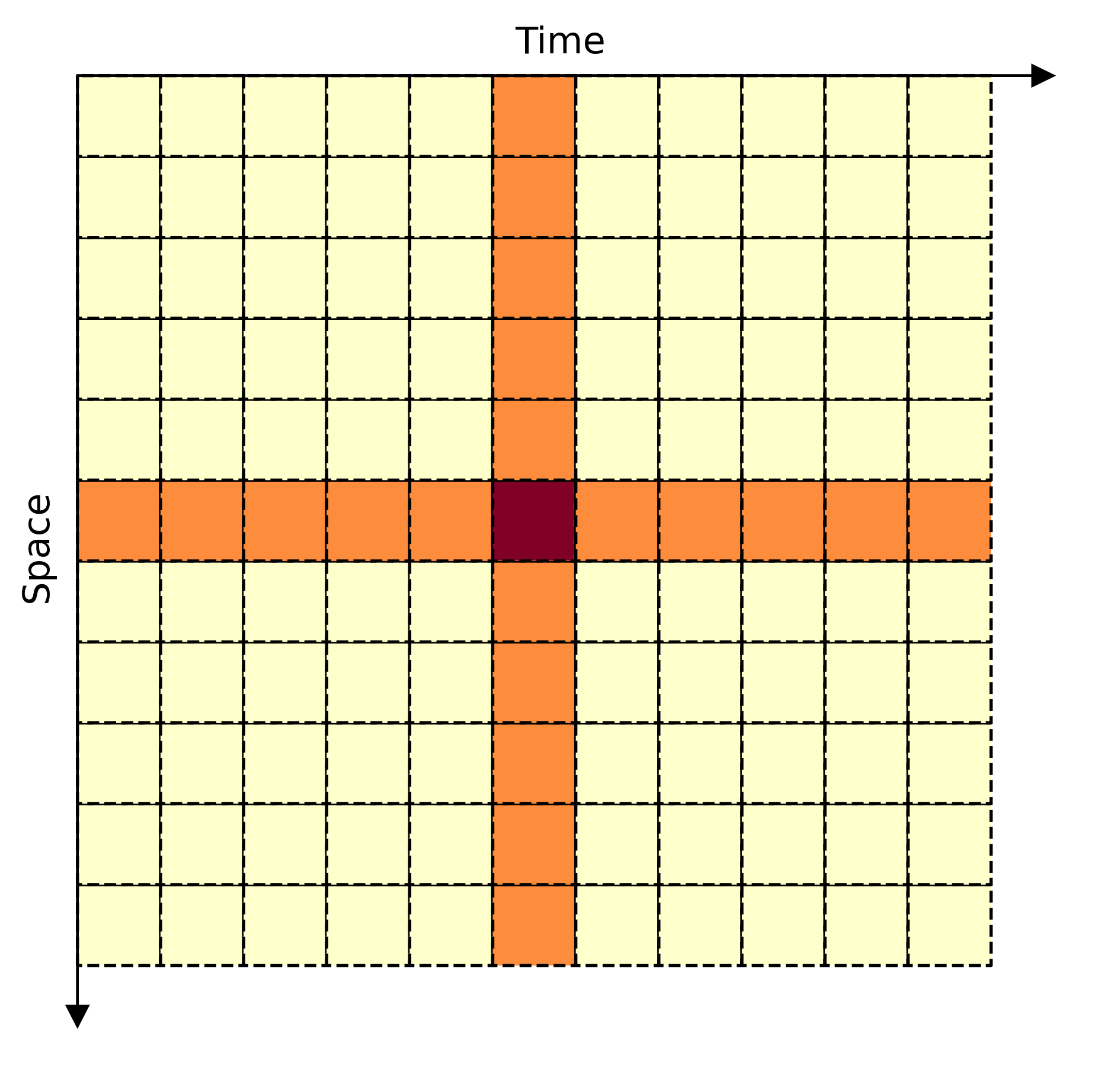}
        \caption{TimeSformer~\citep{bertasius2021space}: $O(T^2S + TS^2)$}
    \end{subfigure} 
    ~
    \begin{subfigure}[t]{0.23\textwidth}
    \centering
        \includegraphics[width=3.3cm]{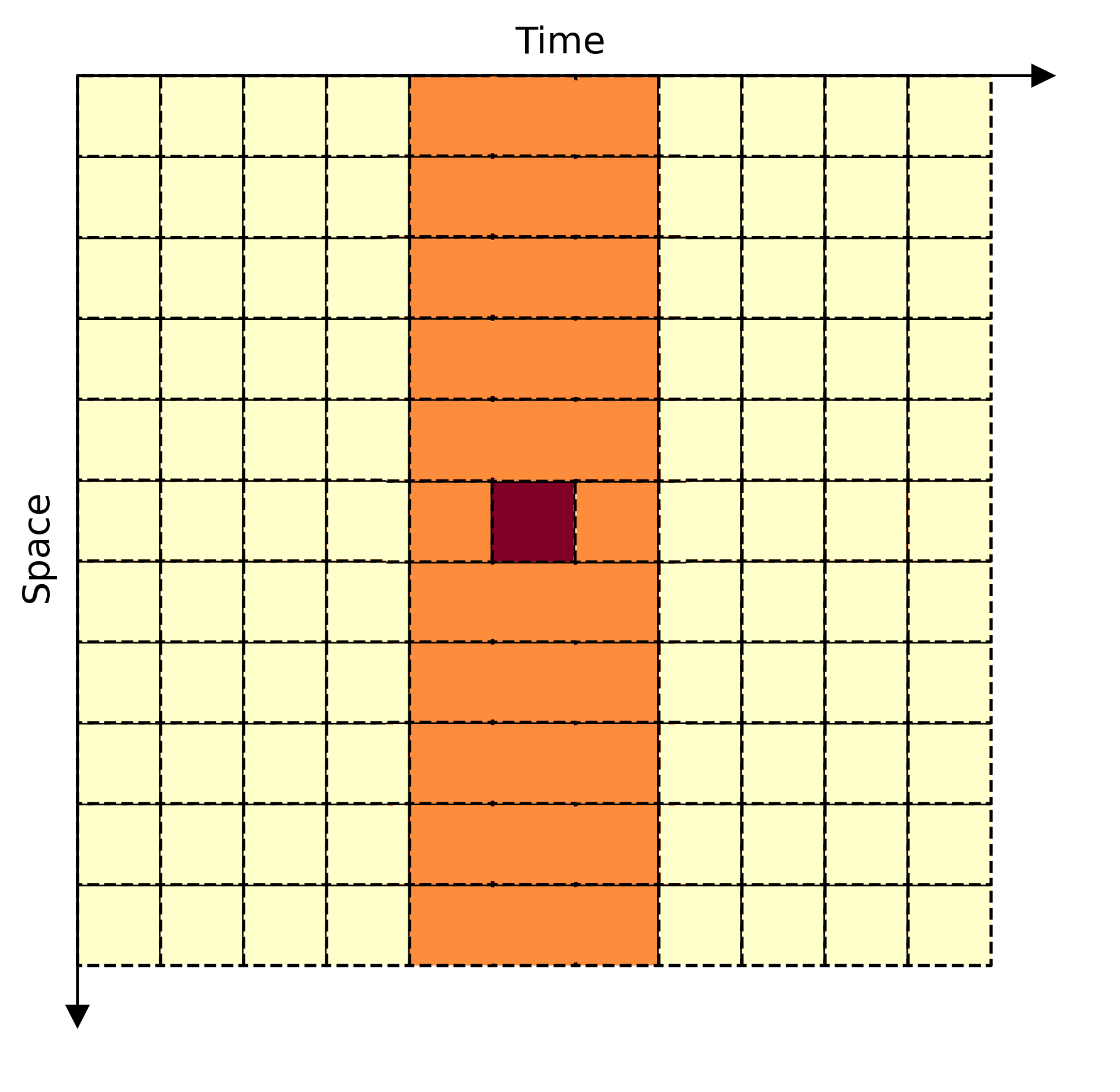}
        \caption{Ours: $O(TS^2)$}
    \end{subfigure}
    \caption{Different approaches to space-time self-attention for video recognition. In all cases, the key locations that the query vector, located at the center of the grid in red, attends are shown in orange. Unlike prior work, our key vector is constructed by mixing information from tokens located at the same spatial location within a local temporal window. Our method then performs self-attention with these tokens. Note that our mechanism allows for an efficient approximation of local space-time attention at no extra cost when compared to a spatial-only attention model.}
    \label{fig:overall_idea}
    \vspace*{-0.4cm}
\end{figure}

\textbf{Our main contribution} in this paper is a Video Transformer model that has complexity $O(TS^2)$ and, hence, is as efficient as the baseline model, yet, as our results show, it outperforms  recently/concurrently proposed work ~\citep{bertasius2021space,arnab2021vivit} in terms of efficiency (\ie accuracy/FLOP) by significant margins. To achieve this, our model makes two approximations to the full space-time attention used in Video Transformers: (a) It restricts time attention to a local temporal window and capitalizes on the Transformer's depth to obtain full temporal coverage of the video sequence. (b) It uses efficient space-time mixing to attend \textit{jointly} spatial and temporal locations without inducing any additional cost on top of a spatial-only attention model. Fig.~\ref{fig:overall_idea} shows the proposed approximation to space-time attention. We also show how to integrate two very lightweight mechanisms for global temporal-only attention, which provide additional accuracy improvements at minimal computational cost. We demonstrate that our model is surprisingly effective in terms of capturing long-term dependencies and producing very high recognition accuracy on the most popular video recognition datasets, including Something-Something-v2~\citep{goyal2017something}, Kinetics~\citep{carreira2017quo} and Epic Kitchens~\citep{damen2018scaling}, while at the same time being significantly more efficient than other Video Transformer models.

\section{Related work}

\textbf{Video recognition:} Standard solutions are based on CNNs and can be broadly classified into two categories: 2D- and 3D-based approaches. 2D-based approaches process each frame independently to extract frame-based features which are then aggregated temporally with some sort of temporal modeling (e.g. temporal averaging) performed at the end of the network ~\citep{wang2018temporal,lin2019tsm,liu2020tam}. The works of~\citep{lin2019tsm,liu2020tam} use the ``shift trick''~\citep{wu2018shift} to have some temporal modeling at a layer level. 3D-based approaches~\citep{carreira2017quo,feichtenhofer2019slowfast,tran2015learning} are considered the current state-of-the-art as they can typically learn stronger temporal models via 3D convolutions. However, they also incur higher computational and memory costs. To alleviate this, a large body of works attempt to improve their efficiency via spatial and/or temporal factorization~\citep{tran2019video,tran2018closer,feichtenhofer2020x3d}.

\textbf{CNN vs ViT:} Historically, video recognition approaches tend to mimic the architectures used for image classification (e.g. from AlexNet~\citep{krizhevsky2012imagenet} to~\citep{karpathy2014large} or from ResNet~\citep{he2016deep} and ResNeXt~\citep{xie2017aggregated} to~\citep{feichtenhofer2019slowfast}). After revolutionizing NLP~\citep{vaswani2017attention,raffel2019exploring}, very recently, Transformer-based architectures showed promising results on large scale image classification too~\citep{dosovitskiy2020image}. While self-attention and attention were previously used in conjunction with CNNs at a layer or block level~\citep{chen20182,zhang2019self,srinivas2021bottleneck}, the Vision Transformer (ViT) of \citet{dosovitskiy2020image} is the first convolution-free, Transformer-based architecture that achieves state-of-the-art results on ImageNet~\citep{deng2009imagenet}.

\textbf{Video Transformer:} Recently/concurrently with our work, vision transformer architectures, derived from~\citep{dosovitskiy2020image}, were used for video recognition~\citep{bertasius2021space,arnab2021vivit}, too. Because performing full space-time attention is computationally prohibitive (\ie $O(T^2S^2)$), their main focus is on reducing this via temporal and spatial factorization. In TimeSformer~\citep{bertasius2021space}, the authors propose  applying spatial and temporal attention in an alternating manner reducing the complexity to $O(T^2S + TS^2)$. In a similar fashion, ViViT~\citep{arnab2021vivit} explores several avenues for space-time factorization. In addition, they also proposed to adapt the patch embedding process from~\citep{dosovitskiy2020image} to 3D (\ie video) data. Our work proposes a completely different approximation to full space-time attention that is also efficient. To this end, we firstly restrict full space-time attention to a local temporal window which is reminiscent of~\citep{beltagy2020longformer} but applied here to space-time attention and video recognition. Secondly, we define a local joint space-time attention which we show that can be implemented efficiently via the ``shift trick''~\citep{wu2018shift}.

\section{Method}\label{sec:method}

\textbf{Video Transformer:} We are given a video clip $\mathbf{X}\in\mathbb{R}^{T\times H \times W \times C }$ ($C=3$, $S=H W$). Following ViT~\citep{dosovitskiy2020image}, each frame is divided into $K\times K$ non-overlapping patches which are then mapped into visual tokens using a linear embedding layer $\mathbf{E}\in\mathbb{R}^{3K^2 \times d}$.  Since self-attention is permutation invariant, in order to preserve the information regarding the location of each patch within space and time, we also learn two positional embeddings, one for space: $\mathbf{p}_{s}\in\mathbb{R}^{1 \times S\times d}$ and one for time: $\mathbf{p}_{t}\in\mathbb{R}^{T\times 1 \times d}$. These are then added to the initial visual tokens. Finally, the token sequence is processed by $L$ Transformer layers.

The visual token at layer $l$, spatial location $s$ and temporal location $t$ is denoted as: 
\begin{equation}
\mathbf{z}^l_{s,t}\in\mathbb{R}^d, \;\;\; l=0,\dots,L-1, \;\; s=0,\dots,S-1, \;\; t=0,\dots,T-1.  
\end{equation}
In addition to the $TS$ visual tokens extracted from the video, a special classification token $\mathbf{z}^l_{cls}\in\mathbb{R}^{d}$ is prepended to the token sequence~\citep{devlin2018bert}.
The $l-$th Transformer layer processes the visual tokens $\mathbf{Z}^l\in\mathbb{R}^{(TS+1) \times d}$ of the previous layer using a series of  Multi-head Self-Attention (MSA), Layer Normalization (LN), and MLP ($\mathbb{R}^d \rightarrow \mathbb{R}^{4d} \rightarrow \mathbb{R}^d$) layers as follows:
\begin{eqnarray}
\mathbf{Y}^{l} & = & \textrm{MSA}(\textrm{LN}(\mathbf{Z}^{l-1})) + \mathbf{Z}^{l-1},\\
\mathbf{Z}^{l+1} & = & \textrm{MLP}(\textrm{LN}(\mathbf{Y}^{l})) + \mathbf{Y}^{l}.
\end{eqnarray}

The main computation of a single full space-time Self-Attention (SA) head boils down to calculating:
\begin{equation}
\mathbf{y}^{l}_{s,t} = \sum_{t'=0}^{T-1} \sum_{s'=0}^{S-1} \textrm{Softmax}\{(\mathbf{q}^{l}_{s,t} \cdot \mathbf{k}^{l}_{s',t'})/\sqrt{d_h}\} \mathbf{v}^{l}_{s',t'}, \;\big\{\begin{smallmatrix}
  s=0,\dots,S-1\\
  t=0,\dots,T-1
\end{smallmatrix}\big\}   
\label{eq:SA}
\end{equation}
where $\mathbf{q}^{l}_{s,t},\mathbf{k}^{l}_{s,t}, \mathbf{v}^{l}_{s,t} \in\mathbb{R}^{d_h}$ are the query, key, and value
vectors computed from $\mathbf{z}^l_{s,t}$ (after LN) using embedding matrices $\mathbf{W_q},\mathbf{W_k}, \mathbf{W_v} \in\mathbb{R}^{d \times d_h}$. Finally, the output of the $h$ heads is concatenated and projected using embedding matrix $\mathbf{W_h}\in\mathbb{R}^{hd_h \times d}$.

The complexity of the full model is: $O(3hTSdd_h)$ ($qkv$ projections) $+ O(2hT^2S^2d_h)$ (MSA for $h$ attention heads) $+ O(TS(hd_h)d)$ (multi-head projection) $+ O(4TSd^2)$ (MLP). From these terms, our goal is to reduce the cost $O(2T^2S^2d_h)$ (for a single attention head) of the full space-time attention which is the dominant term. For clarity, from now on, we will drop constant terms and $d_h$ to report complexity unless necessary. Hence, the complexity of the full space-time attention is $O(T^2S^2)$. 

\textbf{Our baseline} is a model that performs a simple approximation to the full space-time attention by applying, at each Transformer layer, spatial-only attention:
\begin{equation}
\mathbf{y}^{l}_{s,t} = \sum_{s'=0}^{S-1} \textrm{Softmax}\{(\mathbf{q}^{l}_{s,t} \cdot \mathbf{k}^{l}_{s',t})/\sqrt{d_h}\} \mathbf{v}^{l}_{s',t}, 
\;\big\{\begin{smallmatrix}
  s=0,\dots,S-1\\
  t=0,\dots,T-1
\end{smallmatrix}\big\}  
\label{eq:sSA}
\end{equation}
the complexity of which is $O(TS^2)$. Notably, the complexity of the proposed space-time mixing attention is also $O(TS^2)$. Following spatial-only attention, simple temporal averaging is performed on the class tokens $\mathbf{z}_{final} = \frac{1}{T}\sum\limits_{t} \mathbf{z}^{L-1}_{t,cls}$ to obtain a single feature that is fed to the linear classifier.  

\textbf{Recent work} by~\citep{bertasius2021space, arnab2021vivit} has focused on reducing the cost $O(T^2S^2)$ of the full space-time attention of Eq. \ref{eq:SA}. \citet{bertasius2021space} proposed the factorised attention:
\begin{equation}
   \begin{split}
        \tilde{\mathbf{y}}^{l}_{s,t} = \sum_{t'=0}^{T-1} \textrm{Softmax}\{(\mathbf{q}^{l}_{s,t} \cdot \mathbf{k}^{l}_{s,t'})/\sqrt{d_h}\} \mathbf{v}^{l}_{s,t'}, \\
        \mathbf{y}^{l}_{s,t} = \sum_{s'=0}^{S-1} \textrm{Softmax}\{\tilde{\mathbf{q}}^{l}_{s,t} \cdot \tilde{\mathbf{k}}^{l}_{s',t})/\sqrt{d_h}\} \tilde{\mathbf{v}}^{l}_{s',t},
    \end{split}
    \quad
    \begin{split}
         \; \begin{Bmatrix}
          s=0,\dots,S-1\\
          t=0,\dots,T-1
        \end{Bmatrix},
    \end{split}
    \label{eq:fSA}
\end{equation}
where $\tilde{\mathbf{q}}^{l}_{s,t},  \tilde{\mathbf{k}}^{l}_{s',t} \tilde{\mathbf{v}}^{l}_{s',t}$ are new query, key and value vectors calculated from $\tilde{\mathbf{y}}^l_{s,t}$~\footnote{More precisely, Eq. \ref{eq:fSA} holds for $h=1$ heads. For $h>1$, the different heads $\tilde{\mathbf{y}}^{l,h}_{s,t}$ are concatenated and projected to produce $\tilde{\mathbf{y}}^l_{s,t}$.}. The above model reduces complexity to $O(T^2S + TS^2)$. However, temporal attention is performed for a fixed spatial location which is  ineffective when there is camera or object motion and there is spatial misalignment between frames. 

The work of~\citep{arnab2021vivit} is concurrent to ours and proposes the following approximation: $L_s$ Transformer layers perform spatial-only attention as in Eq.~\ref{eq:sSA} (each with complexity $O(S^2)$). Following this, there are $L_t$ Transformer layers performing temporal-only attention on the class tokens $\mathbf{z}^{L_s}_{t}$. The complexity of the temporal-only attention is, in general, $O(T^2)$.

\textbf{Our model} aims to better approximate the full space-time self-attention (SA) of Eq.~\ref{eq:SA} while keeping complexity to $O(TS^2)$, i.e. inducing no further complexity to a spatial-only model. 

To achieve this, we make a first approximation to perform full space-time attention but restricted to a local temporal window $[-t_w, t_w]$:  
\begin{equation}\label{eq:localSA}
\mathbf{y}^{l}_{s,t} = \sum_{t'=t-t_w}^{t+t_w} \sum_{s'=0}^{S-1} \textrm{Softmax}\{(\mathbf{q}^{l}_{s,t} \cdot \mathbf{k}^{l}_{s',t'})/\sqrt{d_h}\} \mathbf{v}^{l}_{s',t'}= \sum_{t'=t-t_w}^{t+t_w} \mathbf{V}^{l}_{t'} \mathbf{a}^l_{t'}, \;\big\{\begin{smallmatrix}
  s=0,\dots,S-1\\
  t=0,\dots,T-1
\end{smallmatrix}\big\}  
\end{equation}
where $\mathbf{V}^{l}_{t'}=[\mathbf{v}^{l}_{0,t'}; \mathbf{v}^{l}_{1,t'}; \dots;  \mathbf{v}^{l}_{S-1,t'}]\in\mathbb{R}^{d_h \times S}$ and $\mathbf{a}^l_{t'}=[a^l_{0,t'}, a^l_{1,t'}, \dots, a^l_{S-1,t'}]\in\mathbb{R}^{S}$ is the vector with the corresponding attention weights. Eq.~\ref{eq:localSA} shows that, for a single Transformer layer, $\mathbf{y}^{l}_{s,t}$ is a spatio-temporal combination of the visual tokens in the local window $[-t_w, t_w]$. It follows that, after $k$ Transformer layers, $\mathbf{y}^{l+k}_{s,t}$ will be a spatio-temporal combination of the visual tokens in the local window $[-kt_w, kt_w]$ which in turn conveniently allows to perform spatio-temporal attention over the whole clip. For example, for $t_w=1$ and $k=4$, the local window becomes $[-4, 4]$ which spans the whole video clip for the typical case $T=8$.

\begin{figure}
    \centering
    \begin{subfigure}[b]{0.4\textwidth}
        \centering
        \includegraphics[trim={0.5cm 0.7cm 0.5cm 0.7cm}, clip,width=5.5cm]{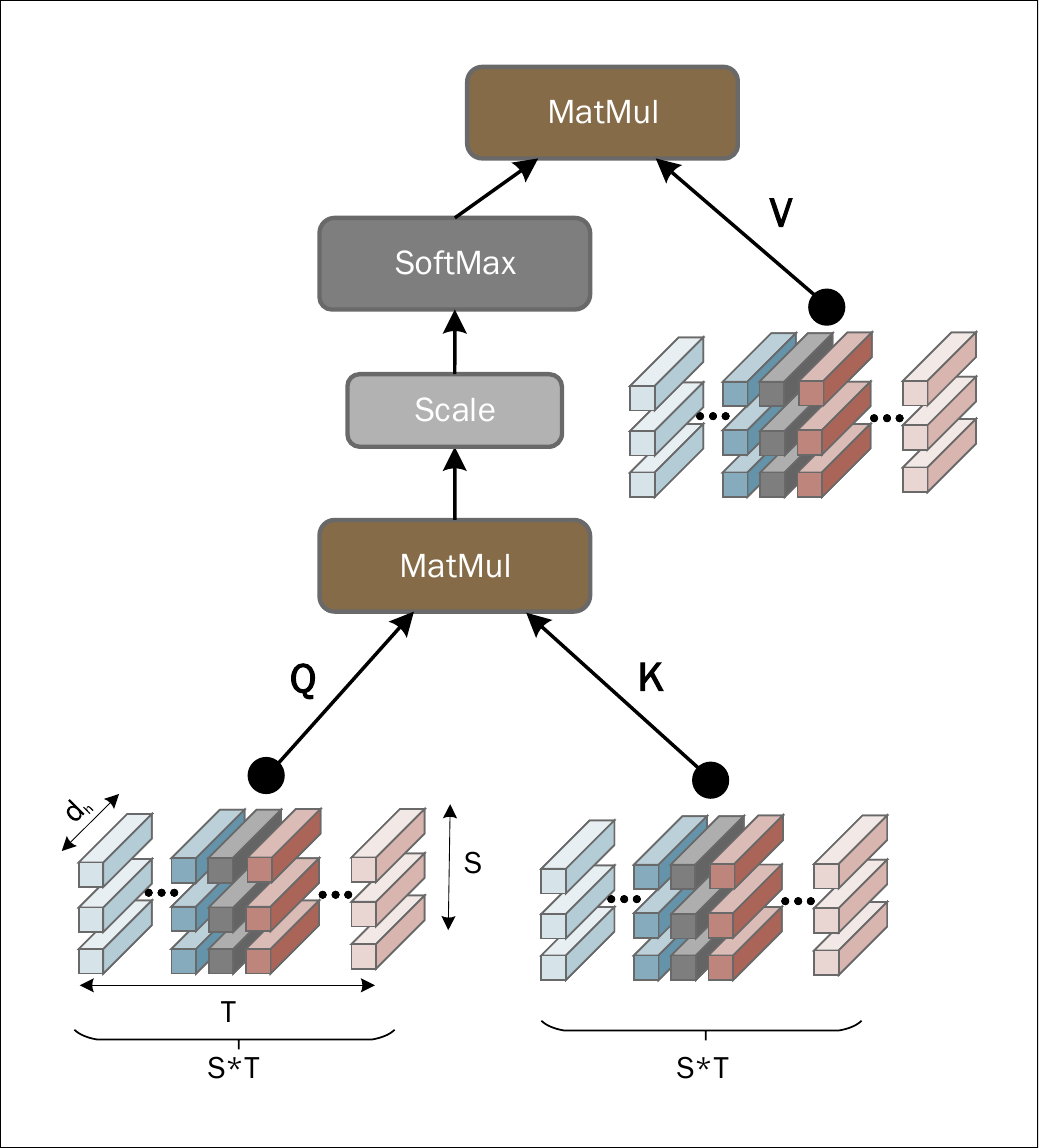}
        \caption{Full Spatio-temporal attention.}
    \end{subfigure}
    ~
    \begin{subfigure}[b]{0.58\textwidth}
        \centering
        \includegraphics[trim={1.8cm 0.7cm 0.5cm 0.7cm}, clip,clip,width=6cm]{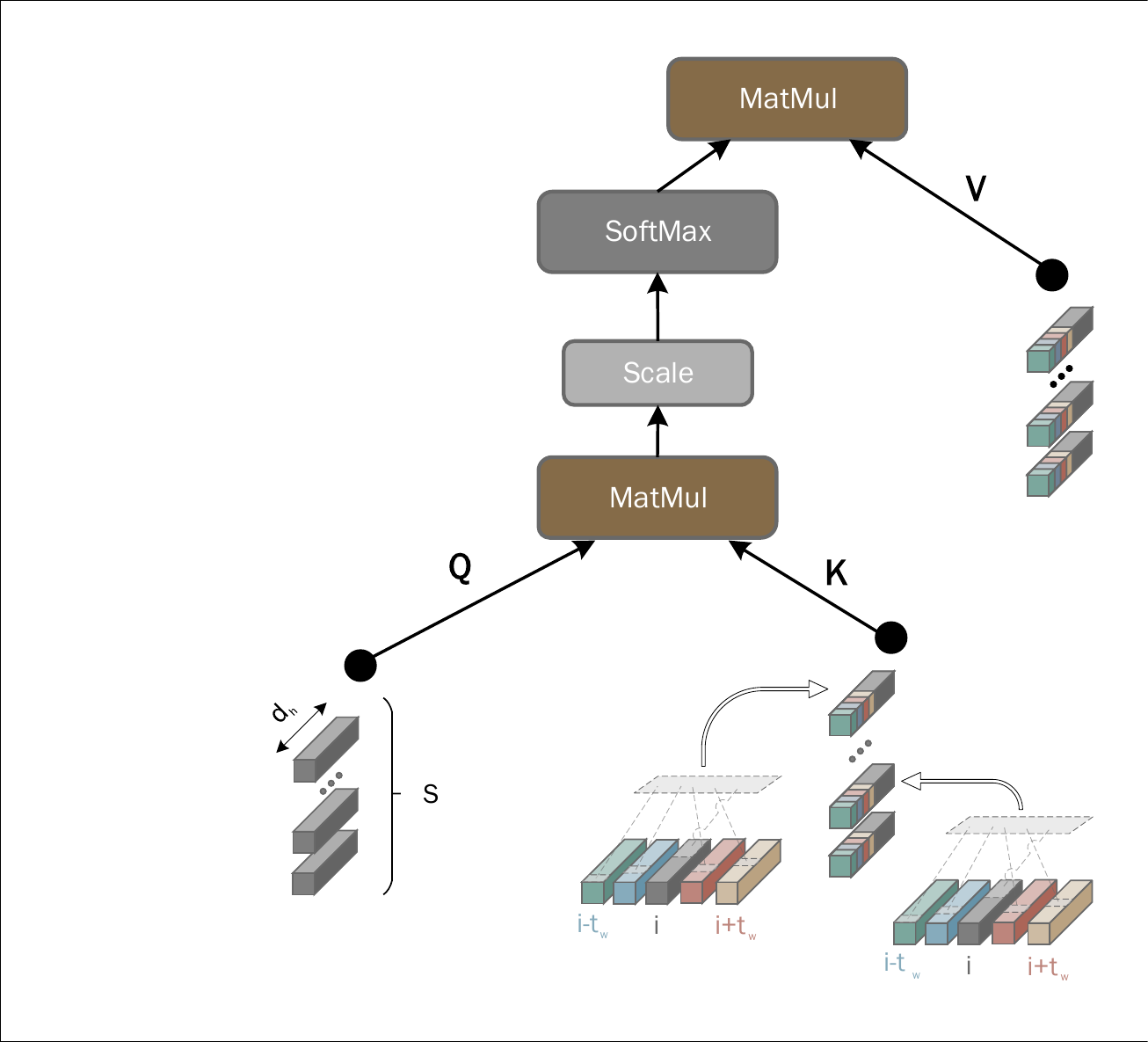}
        \caption{Ours.}
    \end{subfigure}

    \caption{Detailed self-attention computation graph for (a) full space-time attention and (b) the proposed space-time mixing approximation. Notice that in our case only S tokens participate instead of TS. The temporal information is aggregated by indexing channels from adjacent frames. Tokens of identical colors share the same temporal index.}
    \label{fig:detailed-diff}
    \vspace*{-0.5cm}
\end{figure}

The complexity of the local self-attention of Eq.~\ref{eq:localSA} is $O((2t_w+1)TS^2)$. To reduce this even further, we make a second approximation on top of the first one as follows: the attention between spatial locations $s$ and $s'$ according to the model of Eq.~\ref{eq:localSA} is:
\begin{equation} \label{eq:localSA2}
\sum_{t'=t-t_w}^{t+t_w}  \textrm{Softmax}\{(\mathbf{q}^{l}_{s,t} \cdot \mathbf{k}^{l}_{s',t'})/\sqrt{d_h}\} \mathbf{v}^{l}_{s',t'},
\end{equation}
i.e. it requires the calculation of $2t_w+1$ attentions, one per temporal location over $[-t_w, t_w]$. Instead, we propose to calculate a single attention over $[-t_w, t_w]$ which can be achieved by $\mathbf{q}^{l}_{s,t}$ attending $\mathbf{k}^{l}_{s',-t_w:t_w} \triangleq [\mathbf{k}^{l}_{s',t-t_w};\dots;\mathbf{k}^{l}_{s',t+t_w}] \in \mathbb{R}^{(2t_w+1)d_h}$. Note that to match the dimensions of $\mathbf{q}^{l}_{s,t}$ and $\mathbf{k}^{l}_{s',-t_w:t_w}$ a further projection of $\mathbf{k}^{l}_{s',-t_w:t_w}$ to $\mathbb{R}^{d_h}$ is normally required which has complexity $O((2t_w+1)d_h^2)$ and hence compromises the goal of an efficient implementation. To alleviate this, we use the ``shift trick''~\cite{wu2018shift, lin2019tsm} which allows to perform both zero-cost dimensionality reduction, space-time mixing and attention (between $\mathbf{q}^{l}_{s,t}$ and  $\mathbf{k}^{l}_{s',-t_w:t_w}$) in $O(d_h)$. In particular, each $t' \in [-t_w, t_w]$ is assigned $d_h^{t'}$ channels from $d_h$ (i.e. $\sum_{t'} d_h^{t'} = d_h$). Let $\mathbf{k}^{l}_{s',t'}(d_h^{t'})\in \mathbb{R}^{d_h^{t'}}$ denote the operator for indexing the $d_h^{t'}$ channels from $\mathbf{k}^{l}_{s',t'}$. Then, a new key vector is constructed as:
\begin{equation}
    \tilde{\mathbf{k}}^{l}_{s',-t_w:t_w} \triangleq [\mathbf{k}^{l}_{s',t-t_w}(d_h^{t-t_w}), \dots, \mathbf{k}^{l}_{s',t+t_w}(d_h^{t+t_w})]\in \mathbb{R}^{d_h}.
    \label{eq:construct_key}
\end{equation}
Fig.~\ref{fig:detailed-diff} shows how the key vector $\tilde{\mathbf{k}}^{l}_{s',-t_w:t_w}$ is constructed. In a similar way, we also construct a new value vector $\tilde{\mathbf{v}}^{l}_{s',-t_w:t_w}$. Finally, the proposed approximation to the full space-time attention is given by: 
\begin{equation}
\mathbf{y}^{l_s}_{s,t} = \sum_{s'=0}^{S-1} \textrm{Softmax}\{(\mathbf{q}^{l_s}_{s,t} \cdot \tilde{\mathbf{k}}^{l}_{s',-t_w:t_w})/\sqrt{d_h}\} \tilde{\mathbf{v}}^{l}_{s',-t_w:t_w}, 
\;\big\{\begin{smallmatrix}
  s=0,\dots,S-1\\
  t=0,\dots,T-1
\end{smallmatrix}\big\}.  
\label{eq:oursSA}
\end{equation}
This has the complexity of a spatial-only attention ($O(TS^2)$) and hence it is more efficient than previously proposed video transformers~\cite{bertasius2021space, arnab2021vivit}. Our model also provides a better approximation to the full space-time attention and as shown by our results it significantly outperforms~\cite{bertasius2021space, arnab2021vivit}.   

\noindent\textbf{Temporal Attention aggregation:} The final set of the class tokens $\mathbf{z}^{L-1}_{t,cls}, 0 \leq t \leq L-1$  are used to generate the predictions. To this end, we propose to consider the following options: (a) simple temporal averaging  $\mathbf{z}_{final} = \frac{1}{T}\sum_{t} \mathbf{z}^{L-1}_{t,cls}$ as in the case of our baseline. (b) An obvious limitation of temporal averaging is that the output is treated purely as an ensemble of per-frame features and, hence, completely ignores the temporal ordering between them. To address this, we propose to use a lightweight Temporal Attention (TA) mechanism that will attend the $T$ classification tokens. In particular, a $\mathbf{z}_{final}$ token attends the sequence $[\mathbf{z}^{L-1}_{0,cls}, \ldots , \mathbf{z}^{L-1}_{T-1,cls}]$ using a temporal Transformer layer and then fed as input to the classifier. This is akin to the (concurrent) work of~\cite{arnab2021vivit} with the difference being that in our model we found that a single TA layer suffices whereas~\cite{arnab2021vivit} uses $L_t$. A consequence of this is that the complexity of our layer is $O(T)$ vs $O(2(L_t-1)T^2 + T)$ of~\cite{arnab2021vivit}.     

\noindent\textbf{Summary token:} As an alternative to TA, herein, we also propose a simple lightweight mechanism for information exchange between different frames at intermediate layers of the network. Given the set of tokens for each frame $t$, $\mathbf{Z}_{t}^{l-1}\in\mathbb{R}^{(S+1)\times d_h}$ (constructed by concatenating all tokens $\mathbf{z}_{s,t}^{l-1}, s=0,\dots,S$),  we compute a new set of $R$ tokens $\mathbf{Z}^{l}_{r, t} = \phi(\mathbf{Z}_{t}^{l-1})\in\mathbb{R}^{R\times d_h}$ which summarize the frame information and hence are named ``Summary'' tokens. These are, then, appended to the visual tokens of all frames to calculate the keys and values so that the query vectors attend the original keys plus the Summary tokens. 
Herein, we explore the case that $\phi(.)$ performs simple spatial averaging $\mathbf{z}^{l}_{0, t} =  \frac{1}{S}\sum_{s} \mathbf{z}^{l}_{s,t}$ over the tokens of each frame ($R=1$ for this case). Note that, for $R=1$, the extra cost that the Summary token induces is $O(TS)$.

\noindent\textbf{X-ViT:} We call the Video Transformer based on the proposed (a) space-time mixing attention and (b)  lightweight global temporal attention (or summary token) as \textbf{X-ViT}. 

\section{Results}\label{sec:results}

\subsection{Experimental setup}\label{ssec:results-experimental}

\noindent \textbf{Datasets:} We trained and evaluated the proposed models on the following datasets (all datasets are publicly available for research purposes):

\textit{Kinetics-400 and 600}: The Kinetics~\citep{kay2017kinetics} dataset consists of short clips (typically 10 sec long) sampled from YouTube and labeled using 400 and 600 classes, respectively. Due to the removal of certain videos from YouTube, the versions of the datasets used in this paper consist of approximately 261k clips for Kinetics-400 and 457k for Kinetics-600. Note, that these numbers are lower than the original datasets and thus might induce a negative performance bias when compared with prior works.

\textit{Something-Something-v2 (SSv2)}: The SSv2~\citep{goyal2017something} dataset consists of 220,487 short videos, with a length between 2 and 6 seconds that picture humans performing pre-defined basic actions with everyday objects. Since the objects and backgrounds in the videos are consistent across different action classes, this dataset tends to require stronger temporal modeling.  Due to this, we conducted most of our ablation studies on SSv2 to better analyze the importance of the proposed components.

\textit{Epic Kitchens-100 (Epic-100)}: Epic-100 is an egocentric large scale action recognition dataset consisting of more than 90,000 action segments that span across 100 hours of recordings in native environments, capturing daily activities~\citep{damen2020rescaling}. The dataset is labeled using 97 verb and 300 noun classes. The evaluation results are reported using the standard action recognition protocol: the network predicts the ``verb'' and the ``noun'' using two heads. The predictions are then merged to construct an ``action'' which is used to calculate the accuracy.

\noindent \textbf{Network architecture:} The backbone models closely follow the ViT architecture~\citep{dosovitskiy2020image}. Most of the experiments were performed using the ViT-B/16 variant ($L=12$, $h=12$, $d=768$, $K=16$), where $L$ represents the number of transformer layers, $h$ the number of heads, $d$ the embedding dimension and $K$ the patch size. We initialized our models from a pretrained ImageNet-21k~\citep{deng2009imagenet} ViT model. The spatial positional encoding $\mathbf{p}_s$ was initialized from the pretrained 2D model and the temporal one, $\mathbf{p}_t$, with zeros so that it does not have a great impact on the tokens early on during training. The models were trained on 8 V100 GPUs using PyTorch~\citep{paszke2019pytorch}.

\textbf{Testing details:} Unless otherwise stated, we used ViT-B/16 and $T=8$ frames. We mostly used Temporal Attention (TA) for temporal aggregation. We report accuracy results for $1\times3$ views (1 temporal clip and 3 spatial crops) departing from the common approach of using up to $10\times3$ views~\citep{lin2019tsm,feichtenhofer2019slowfast}. The $1\times3$ views setting was also used in~\citet{bertasius2021space}. To measure the variation between runs, we trained one of the 8--frame models 5 times. The results varied by $\pm 0.4\%$. 

\subsection{Ablation studies}\label{ssec:results-ablation}

Throughout this section we study the effect of varying certain design choices and different components of our method. Because SSv2 tends to require a more fine-grained temporal modeling, unless otherwise specified, all results reported, in this subsection, are on the SSv2. 

\begin{wraptable}[14]{r}{5.5cm}
\vspace{-0.4cm}
    \caption{Effect of local window size. To isolate its effect from that of temporal aggregation, the models were trained using temporal averaging. Note, that \textit{(Bo.)} indicates that only features from the boundaries of the local window were used, ignoring the intermediate ones. }\label{tab:local-window-size}
    
    \centering
    \begin{tabular}{ccc}
        \toprule
      Variant & Top-1 & Top-5  \\
       \midrule
        $t_w=0$ & 45.2 & 71.4  \\
        $t_w=1$ & \textbf{62.5} & \textbf{87.8}  \\
        $t_w=2$ & 60.5 & 86.4  \\
        $t_w=2$ (Bo.) & 60.4 & 86.2  \\
        \bottomrule
    \end{tabular}
\end{wraptable}

\noindent\textbf{Effect of local window size:} Table~\ref{tab:local-window-size} shows the accuracy of our model by varying the local window size $[-t_w, t_w]$ used in the proposed space-time mixing attention. Firstly, we observe that the proposed model is significantly superior to our baseline ($t_w=0$) which uses spatial-only attention. Secondly, a window of $t_w=1$ produces the best results. This shows that more gradual increase of the effective window size that is attended is more beneficial compared to more aggressive ones, i.e. the case where $t_w=2$. A performance degradation for the case $t_w=2$ could be attributed to boundary effects (handled by filling with zeros) which are aggravated as $t_w$ increases. Based on these results, we chose to use $t_w=1$ for the models reported hereafter.

\noindent\textbf{Effect of SA position:} We explored which layers should the proposed space-time mixing attention operation be applied to \textit{within the Transformer}. Specifically, we explored the following variants: Applying it to the first $L/2$ layers, to the last $L/2$ layers, to every odd indexed layer and finally, to all layers. As the results from Table~\ref{tab:SSv2-mixinglocation} show, the exact layers within the network that self-attention is applied to do not matter; what matters is the number of layers it is applied to. We attribute this result to the increased temporal receptive field and cross-frame interactions.

\noindent\textbf{Effect of temporal aggregation:} Herein, we compare the two methods used for temporal aggregation: simple temporal averaging~\citep{wang2016temporal} and the proposed Temporal Attention (TA) mechanism. Given that our model already incorporates temporal information through the proposed space-time attention, we also explored how many TA layers are needed. As shown in Table~\ref{tab:SSv2-vit-num-layers} replacing temporal averaging with one TA layer improves the Top-1 accuracy from 62.5\% to 64.4\%. Increasing the number of layers further yields no additional benefits. We also report the accuracy of spatial-only attention plus TA aggregation. In the absence of the proposed space-time mixing attention, the TA layer alone is unable to compensate, scoring only 56.6\% as shown in Table~\ref{tab:SSv2-mixingamount}. This highlights the need of having both components in our final model. For the next two ablation studies, we therefore used 1 TA layer.

\begin{table}[t]
    \centering
    \caption{Effect of (a) proposed SA position, (b) temporal aggregation and number of Temporal Attention (TA) layers, (c) space-time mixing $qkv$ vectors and (d) amount of mixed channels on SSv2.}
    \label{tab:ablations}
    \begin{subtable}[t]{.5\linewidth}
        \caption{Effect of applying the proposed SA to certain layers.}\label{tab:SSv2-mixinglocation}
    \centering
    \begin{tabular}{ccc}
        \toprule
       Transform. layers & Top-1 & Top-5 \\
       \midrule
       1st half & 61.7 & 86.5 \\
       2nd half & 61.6 & 86.3 \\
       Half (odd. pos) & 61.2 & 86.4 \\
       All & \textbf{62.6} & \textbf{87.8} \\
        \bottomrule
    \end{tabular}
        \caption{Effect of number of TA layers. 0 corresponds to temporal averaging.}\label{tab:SSv2-vit-num-layers}
    \centering
    \begin{tabular}{ccc}
        \toprule
        \#. TA layers & Top-1 & Top-5 \\
       \midrule
       0 (temp. avg.) & 62.4 & 87.8 \\
       1 & 64.4 & \textbf{89.3} \\
       2 & \textbf{64.5} & \textbf{89.3} \\
       3 & \textbf{64.5} & \textbf{89.3} \\
        \bottomrule
    \end{tabular}
    \end{subtable}
    \hfill
      \begin{subtable}[t]{.45\linewidth}
        \caption{Effect of space-time mixing. x denotes the input token before $qkv$ projection. Query produces equivalent results with key and thus omitted.}\label{tab:SSv2-mixingplacemnt}
    \centering
    \begin{tabular*}{\textwidth}{c@{\extracolsep{\fill}}cccc}
        \toprule
       x & key & value & Top-1 & Top-5 \\
       \midrule
       \xmark & \xmark & \xmark & 56.6 & 83.5 \\
       \checkmark & \xmark & \xmark & 63.1 & 88.8 \\
       \xmark & \checkmark & \xmark & 63.1 & 88.8 \\
       \xmark & \xmark & \checkmark & 62.5 & 88.6 \\
       \xmark & \checkmark & \checkmark & \textbf{64.4} & \textbf{89.3} \\
        \bottomrule
    \end{tabular*}
        \caption{Effect of amount of mixed channels. * uses temp. avg. aggregation.}\label{tab:SSv2-mixingamount}
    
    \centering
    \begin{tabular*}{\textwidth}{c@{\extracolsep{\fill}}cccc}
        \toprule
      0\%* & 0\% & 25\% & 50\% & 100\% \\
       \midrule
       45.2 & 56.6 & 64.3  & \textbf{64.4} & 62.5 \\
        \bottomrule
    \end{tabular*}
    \end{subtable}
\vspace*{-0.3cm}
\end{table}

\begin{table}[t]
    \centering
    \caption{Comparison between TA and Summary token on SSv2 (left) and Kinetics-400 (right).}
    \label{tab:ablations-summarization}
    \begin{subtable}[t]{.47\linewidth}
            \centering
            \begin{tabular}{cccc}
                \toprule
               Summary & TA & Top-1 & Top-5 \\
               \midrule
               \xmark & \xmark & 62.4 & 87.8 \\
               \checkmark & \xmark & 63.7 & 88.9 \\
               \checkmark & \checkmark & 63.4 & 88.9 \\
               \xmark & \checkmark & \textbf{64.4} & \textbf{89.3} \\
                \bottomrule
            \end{tabular}
    \end{subtable}
    \hfill
    \begin{subtable}[t]{.47\linewidth}
    \centering
            \begin{tabular}{cccc}
                \toprule
               Summary & TA & Top-1 & Top-5 \\
               \midrule
               \xmark & \xmark & 77.8 & 93.7 \\
               \checkmark & \xmark & \textbf{78.7} & \textbf{93.7} \\
               \checkmark & \checkmark & 78.0 & 93.2 \\
               \xmark & \checkmark & 78.5 & \textbf{93.7} \\
                \bottomrule
            \end{tabular}
    \end{subtable}
\vspace*{-0.4cm}
\end{table}

\begin{table}[t]
\caption{Comparison with state-of-the-art on the Kinetics-600 dataset. $T\times$ is the number of frames used by our method.}\label{tab:sota-k600}
\centering
    \begin{tabular}{ccccc}
        \toprule
      Method &  Top-1 & Top-5 & Views & FLOPs ($\times 10^9$) \\
      \midrule
      AttentionNAS~\citep{wang2020attentionnas} & 79.8 & 94.4 & - & 1,034 \\
      LGD-3D R101~\citep{qiu2019learning} & 81.5 & 95.6 & $10\times 3$ & - \\
      SlowFast R101+NL~\citep{feichtenhofer2019slowfast} & 81.8 & 95.1 & $10 \times 3$ &  3,480  \\
      X3D-XL~\citep{feichtenhofer2020x3d} & 81.9 & 95.5 & $10\times3$ & 1,452  \\
      TimeSformer-HR~\citep{bertasius2021space} & 82.4 & 96.0 & $1\times 3$ & 5,110  \\
      ViViT-L/16x2~\citep{arnab2021vivit} & 82.5 & 95.6 & $4\times3$ & 17,352  \\
      \midrule
      X-ViT (8$\times$) (Ours) & 82.5 & 95.4 & $1\times 3$ & 425 \\
      X-ViT (16$\times$) (Ours) & \textbf{84.5} & \textbf{96.3} & $1\times 3$ & 850 \\
        \bottomrule
    \end{tabular}
\end{table}

\begin{wraptable}[9]{r}{5.5cm}
    \vspace{-0.4cm}
    \caption{Effect of number of tokens on SSv2.}\label{tab:SSv2-arch-variations}
    
    \centering
    \begin{tabular}{ccccc}
        \toprule
      Variant & Top-1 & Top-5 &  \begin{tabular}{@{}c@{}}FLOPs \\ ($\times 10^9$)\end{tabular}  \\
       \midrule
        ViT-B/32 & 60.5 & 87.4 & 95 \\
        ViT-L/32 & 61.8 & 88.3 & 327 \\
       ViT-B/16 & \textbf{64.4} & \textbf{89.3} & 425 \\
        \bottomrule
    \end{tabular}

\end{wraptable}

\noindent\textbf{Effect of space-time mixing $qkv$ vectors:} Paramount to our work is the proposed space-time mixing attention of Eq.~\ref{eq:oursSA} which is implemented by constructing $\tilde{\mathbf{k}}_{s',-t_w:t_w}^l$ and $\tilde{\mathbf{v}}_{s',-t_w:t_w}^l$ efficiently via channel indexing (see Eq.~\ref{eq:construct_key}). Space-time mixing though can be applied in several different ways in the model. For completeness, herein, we study the effect of space-time mixing to various combinations for the key, value and to the input token prior to $qkv$ projection. As shown in Table~\ref{tab:SSv2-mixingplacemnt}, the combination corresponding to our model (\ie space-time mixing applied to the key and value) significantly outperforms all other variants by up to 2\%. This result is important as it confirms that our model, derived from the proposed approximation to the local space-time attention, gives the best results when compared to other non-well motivated variants.

\noindent\textbf{Effect of amount of space-time mixing:} We define as $\rho d_h$ the total number of channels imported from the adjacent frames in the local temporal window $[-t_w,t_w]$ (\ie $\sum_{t'=-t_w, t\neq 0}^{t_w} d_h^{t'} = \rho d_h$) when constructing $\tilde{\mathbf{k}}_{s',-t_w:t_w}^l$ (see  Section~\ref{sec:method}). Herein, we study the effect of $\rho$ on the model's accuracy. As the results from Table~\ref{tab:SSv2-mixingamount} show, the optimal $\rho$ is between 25\% and 50\%. Increasing $\rho$ to 100\% (\ie all channels are coming from adjacent frames) unsurprisingly degrades the performance as it excludes the case $t'=t$ when performing the self-attention.

\noindent\textbf{Effect of Summary token:} Herein, we compare Temporal Attention with Summary token on SSv2 and Kinetics-400. We used both datasets for this case as they require different type of understanding: fine-grained temporal (SSv2) and spatial content (K400). From Table~\ref{tab:ablations-summarization}, we conclude that the Summary token compares favorable on Kinetics-400 but not on SSv2 showing that is more useful in terms of capturing spatial information. Since the improvement is small, we conclude that 1 TA layer is the best global attention-based mechanism for improving the accuracy of our method adding also negligible computational cost. 

\noindent\textbf{Effect of the number of input frames:} Herein, we evaluate the impact of increasing the number of input frames $T$ from 8 to 16 and 32. We note that, for our method, this change results in a linear increase in complexity. As the results from Table~\ref{tab:SSv2-sota} show, increasing the number of frames from 8 to 16 offers a 1.8\% boost in Top-1 accuracy on SSv2. Moreover, increasing the number of frames to 32 improves the performance by a further 0.2\%, offering diminishing returns. Similar behavior can be observed on Kinetics and Epic-100 in Tables~\ref{tab:k400-sota} and~\ref{tab:sota-epic}. 

\noindent\textbf{Effect of number of tokens:} Herein, we vary the number of input tokens by changing the patch size $K$. As the results from Table~\ref{tab:SSv2-arch-variations} show, even when the number of tokens decreases significantly (ViT-B/32) our approach is still able to produce models that achieve satisfactory accuracy. The benefit of that is having a model which is significantly more efficient. 

\noindent \textbf{Effect of the number of crops at test time.} Throughout this work, at test time, we reported results using 1 temporal and 3 spatial crops (\ie $1\times 3$). This is noticeable different from the current practice of using up to $10\times3$ crops~\citep{feichtenhofer2019slowfast,arnab2021vivit}. 

\begin{figure}[h]
    \centering
    \includegraphics[width=7.5cm]{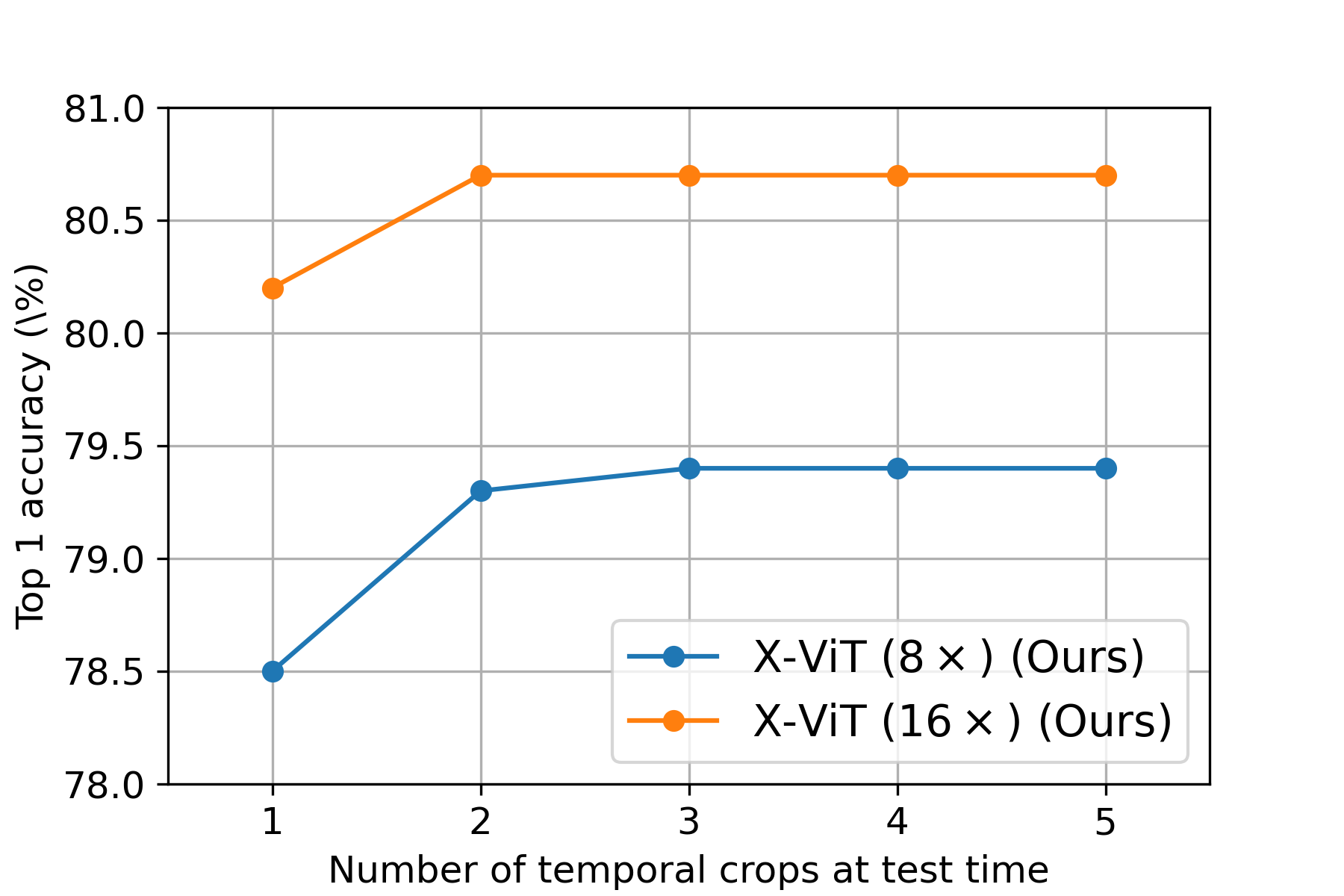}
    \caption{Effect of the number of temporal crops at test time as measured on Kinetics 400 in terms of Top 1 accuracy. For each temporal crop, 3 spatial clips are sampled, for a total of $t_{crops} \times 3$ clips. Notice that beyond $t_{crops}=2$ no additional accuracy gains are observed.}
    \label{fig:k400-ncrops}
\end{figure}

To showcase the behavior of our method, herein, we test the effect of increasing the number of crops on Kinetics-400. As the results from Fig.~\ref{fig:k400-ncrops} show, increasing the number of crops beyond two temporal views (\ie $2\times 3$), yields no additional gains. Our findings align with the ones from the work of~\citet{bertasius2021space} that observes the same properties for the transformer-based architectures.

\noindent \textbf{Latency and throughput considerations:} While the channel shifting operation used by the proposed space-time mixing attention is zero-FLOP, there is still a small cost associated with memory movement operations. In order to ascertain that the induced cost does not introduce noticeable performance degradation, we benchmarked a Vit-B/16 (8$\times$ frames) model using spatial-only attention and the proposed one on 8 V100 GPUs and a batch size of 128. The spatial-only attention model has a throughput of 312 frames/second while our model 304 frames/second.

\begin{table}[ht!]
    \caption{Comparison with state-of-the-art on the Kinetics-400. $T\times$ is the number of frames used by our method.}\label{tab:k400-sota}
    
    \centering
    \begin{tabular}{ccccc}
        \toprule
       Method &  Top-1 & Top-5 & Views & FLOPs ($\times 10^9$) \\
       \midrule
       bLVNet~\citep{fan2019more} & 73.5 & 91.2 & 3 $\times$ 3 & 840\\
       STM~\citep{jiang2019stm} & 73.7 & 91.6 & - & - \\
       TEA~\citep{li2020tea} & 76.1 & 92.5 & 10 $\times$ 3 & 2,100 \\
       TSM R50~\citep{lin2019tsm} & 74.7 & - & 10 $\times$ 3 & 650 \\
       I3D NL~\citep{wang2018non} & 77.7 & 93.3 & 10 $\times$ 3 &  10,800 \\
       CorrNet-101~\citep{wang2020video} & 79.2 & - & 10 $\times$ 3 &  6,700 \\
       ip-CSN-152~\citep{tran2019video} & 79.2 & 93.8 & 10 $\times$ 3 &  3,270 \\
       LGD-3D R101~\citep{qiu2019learning} & 79.4 & 94.4 & - & - \\
       SlowFast 8$\times$8 R101+NL~\citep{feichtenhofer2019slowfast} & 78.7 & 93.5 & 10 $\times$ 3 &  3,480 \\
       SlowFast 16$\times$8 R101+NL~\citep{feichtenhofer2019slowfast} & 79.8 & 93.9 & 10 $\times$ 3 &  7,020 \\
       X3D-XXL~\citep{feichtenhofer2020x3d} & 80.4 & 94.6 & 10 $\times$ 3 &  5,823 \\
       TimeSformer-L~\citep{bertasius2021space} & \textbf{80.7} & 94.7 & 1 $\times$ 3 & 7,140 \\
       ViViT-L/16x2~\citep{bertasius2021space} & 80.6 & 94.7 & 4 $\times$ 3 & 17,352 \\
       \midrule
       X-ViT (8$\times$) (Ours) & 78.5 & 93.7 & 1 $\times$ 3 & 425 \\
       X-ViT (16$\times$) (Ours) & 80.2 & \textbf{94.7} & 1 $\times$ 3 & 850 \\
        \bottomrule
    \end{tabular}

\end{table}

\begin{table}[ht!]
    \caption{Comparison with state-of-the-art on SSv2. $T\times$ is the number of frames used by our method. * - denotes models pretrained on Kinetics-600.}\label{tab:SSv2-sota}
    
    \centering
    \begin{tabular}{ccccc}
        \toprule
       Method &  Top-1 & Top-5 & Views & FLOPs ($\times 10^9$) \\
       \midrule
       TRN~\citep{zhou2018temporal} & 48.8 & 77.6 & - & - \\
       SlowFast+multigrid~\citep{wu2020multigrid} & 61.7 & - &  $1\times 3$ & - \\
       TimeSformer-L~\citep{bertasius2021space} & 62.4 & - & 1 $\times$ 3 & 7,140  \\
       TSM R50~\citep{lin2019tsm} & 63.3 & 88.5 & 2 $\times$ 3 & - \\
       STM~\citep{jiang2019stm} & 64.2 & 89.8 & - & - \\
       MSNet~\citep{kwon2020motionsqueeze} & 64.7 & 89.4 & - & - \\
       TEA~\citep{li2020tea} & 65.1 & - & - & -\\
       ViViT-L/16x2~\citep{bertasius2021space} & 65.4 & 89.8 & 4 $\times$ 3 & 11,892 \\
       \midrule
       X-ViT (8$\times$) (Ours) & 64.4 & 89.3 & 1 $\times$ 3 & 425 \\
       X-ViT (16$\times$) (Ours) & 66.2 & 90.6 & 1 $\times$ 3 & 850 \\
       X-ViT* (16$\times$) (Ours) & \textbf{67.2} & \textbf{90.8} & 1 $\times$ 3 & 850 \\
       X-ViT (32$\times$) (Ours) & 66.4 & 90.7 & 1 $\times$ 3 & 1,270 \\
        \bottomrule
    \end{tabular}
    %\vspace*{-0.4cm}
\end{table}

\subsection{Comparison to state-of-the-art}\label{ssec:results-sota}

Our best model uses the proposed space-time mixing attention in all the Transformer layers and performs temporal aggregation using a single lightweight temporal transformer layer as described in Section~\ref{sec:method}. Unless otherwise specified, we report the results using the $1 \times 3$ configuration for the views (1 temporal and 3 spatial) for all datasets.

On \textbf{Kinetics-400},  we match the current state-of-the-art results while being significantly faster than the next two best recently/concurrently proposed methods that also use Transformer-based architectures: $20\times$ faster than ViVit~\citep{arnab2021vivit} and $8\times$ than TimeSformer-L~\citep{bertasius2021space}. Note that both models from~\citep{arnab2021vivit,bertasius2021space} and ours were initialized from a ViT model pretrained on ImageNet-21k~\citep{deng2009imagenet} and take as input frames at a resolution of $224\times 224$px. Similarly, on Kinetics-600 we set a new state-of-the-art result. See Table~\ref{tab:sota-k600}.

On \textbf{SSv2} we match and surpass the current state-of-the-art, especially in terms of Top-5 accuracy (ours: 90.8\% vs ViViT: 89.8\%~\citep{arnab2021vivit}) using models that are $14\times$ (16 frames) and $9 \times$ (32 frames) faster.

Finally, we observe similar outcomes on \textbf{Epic-100} where we set a new state-of-the-art, showing particularly large improvements especially for ``Verb'' accuracy, while again being more efficient.

\begin{wraptable}[15]{r}{7cm}
\vspace{-0.5cm}
\centering
\caption{Comparison with state-of-the-art on Epic-100. $T\times$ is the \#frames used by our method. Results for other methods are taken from~\citep{arnab2021vivit}.}\label{tab:sota-epic}
    \begin{tabular}{cccc}
        \toprule
      Method &  Action & Verb & Noun  \\
      \midrule
      TSN~\citep{wang2016temporal} & 33.2 & 60.2 & 46.0 \\
      TRN~\citep{zhou2018temporal} & 35.3 & 65.9 & 45.4 \\
      TBN~\citep{kazakos2019epic} & 36.7 & 66.0 & 47.2 \\
      TSM~\citep{kazakos2019epic} & 38.3 & 67.9 & 49.0 \\
      SlowFast~\citep{feichtenhofer2019slowfast} & 38.5 & 65.6 & 50.0  \\
      ViViT-L/16x2~\citep{arnab2021vivit} & 44.0 & 66.4 & \textbf{56.8} \\
      \midrule
      X-ViT (8$\times$) (Ours) & 41.5 & 66.7 & 53.3  \\
      X-ViT (16$\times$) (Ours) & \textbf{44.3} & \textbf{68.7} & 56.4 \\
        \bottomrule
    \end{tabular}
\end{wraptable}

\section{Conclusions}\label{sec:conclusions}

We presented a novel approximation to the full space-time attention that is amenable to an efficient implementation and applied it to video recognition. Our approximation has the same computational cost as spatial-only attention yet the resulting Video Transformer model was shown to be significantly more efficient than recently/concurrently proposed Video Transformers~\cite{bertasius2021space,arnab2021vivit}. By no means this paper proposes a complete solution to video recognition using Video Transformers. Future efforts could include combining our approaches with other architectures than the standard ViT, removing the dependency on pre-trained models and applying the model to other video-related tasks like detection and segmentation. Finally, further research is required for deploying our models on low power devices.

% \clearpage

\medskip

{
\small
\bibliographystyle{plainnat}
\bibliography{egbib}

\begin{thebibliography}{45}
\providecommand{\natexlab}[1]{#1}
\providecommand{\url}[1]{\texttt{#1}}
\expandafter\ifx\csname urlstyle\endcsname\relax
  \providecommand{\doi}[1]{doi: #1}\else
  \providecommand{\doi}{doi: \begingroup \urlstyle{rm}\Url}\fi

\bibitem[Arnab et~al.(2021)Arnab, Dehghani, Heigold, Sun, Lu{\v{c}}i{\'c}, and
  Schmid]{arnab2021vivit}
Anurag Arnab, Mostafa Dehghani, Georg Heigold, Chen Sun, Mario Lu{\v{c}}i{\'c},
  and Cordelia Schmid.
\newblock Vivit: A video vision transformer.
\newblock \emph{arXiv preprint arXiv:2103.15691}, 2021.

\bibitem[Beltagy et~al.(2020)Beltagy, Peters, and Cohan]{beltagy2020longformer}
Iz~Beltagy, Matthew~E Peters, and Arman Cohan.
\newblock Longformer: The long-document transformer.
\newblock \emph{arXiv preprint arXiv:2004.05150}, 2020.

\bibitem[Bertasius et~al.(2021)Bertasius, Wang, and
  Torresani]{bertasius2021space}
Gedas Bertasius, Heng Wang, and Lorenzo Torresani.
\newblock Is space-time attention all you need for video understanding?
\newblock \emph{arXiv preprint arXiv:2102.05095}, 2021.

\bibitem[Carreira and Zisserman(2017)]{carreira2017quo}
Joao Carreira and Andrew Zisserman.
\newblock Quo vadis, action recognition? a new model and the kinetics dataset.
\newblock In \emph{proceedings of the IEEE Conference on Computer Vision and
  Pattern Recognition}, pages 6299--6308, 2017.

\bibitem[Chen et~al.(2018{\natexlab{a}})Chen, Firat, Bapna, Johnson, Macherey,
  Foster, Jones, Parmar, Schuster, Chen, et~al.]{chen2018best}
Mia~Xu Chen, Orhan Firat, Ankur Bapna, Melvin Johnson, Wolfgang Macherey,
  George Foster, Llion Jones, Niki Parmar, Mike Schuster, Zhifeng Chen, et~al.
\newblock The best of both worlds: Combining recent advances in neural machine
  translation.
\newblock \emph{arXiv preprint arXiv:1804.09849}, 2018{\natexlab{a}}.

\bibitem[Chen et~al.(2018{\natexlab{b}})Chen, Kalantidis, Li, Yan, and
  Feng]{chen20182}
Yunpeng Chen, Yannis Kalantidis, Jianshu Li, Shuicheng Yan, and Jiashi Feng.
\newblock A2-nets: Double attention networks.
\newblock \emph{arXiv preprint arXiv:1810.11579}, 2018{\natexlab{b}}.

\bibitem[Damen et~al.(2018)Damen, Doughty, Farinella, Fidler, Furnari, Kazakos,
  Moltisanti, Munro, Perrett, Price, et~al.]{damen2018scaling}
Dima Damen, Hazel Doughty, Giovanni~Maria Farinella, Sanja Fidler, Antonino
  Furnari, Evangelos Kazakos, Davide Moltisanti, Jonathan Munro, Toby Perrett,
  Will Price, et~al.
\newblock Scaling egocentric vision: The epic-kitchens dataset.
\newblock In \emph{ECCV}, 2018.

\bibitem[Damen et~al.(2020)Damen, Doughty, Farinella, Furnari, Kazakos, Ma,
  Moltisanti, Munro, Perrett, Price, et~al.]{damen2020rescaling}
Dima Damen, Hazel Doughty, Giovanni~Maria Farinella, Antonino Furnari,
  Evangelos Kazakos, Jian Ma, Davide Moltisanti, Jonathan Munro, Toby Perrett,
  Will Price, et~al.
\newblock Rescaling egocentric vision.
\newblock \emph{arXiv preprint arXiv:2006.13256}, 2020.

\bibitem[Deng et~al.(2009)Deng, Dong, Socher, Li, Li, and
  Fei-Fei]{deng2009imagenet}
Jia Deng, Wei Dong, Richard Socher, Li-Jia Li, Kai Li, and Li~Fei-Fei.
\newblock Imagenet: A large-scale hierarchical image database.
\newblock In \emph{2009 IEEE conference on computer vision and pattern
  recognition}, pages 248--255. Ieee, 2009.

\bibitem[Devlin et~al.(2018)Devlin, Chang, Lee, and Toutanova]{devlin2018bert}
Jacob Devlin, Ming-Wei Chang, Kenton Lee, and Kristina Toutanova.
\newblock Bert: Pre-training of deep bidirectional transformers for language
  understanding.
\newblock \emph{arXiv preprint arXiv:1810.04805}, 2018.

\bibitem[Dosovitskiy et~al.(2020)Dosovitskiy, Beyer, Kolesnikov, Weissenborn,
  Zhai, Unterthiner, Dehghani, Minderer, Heigold, Gelly,
  et~al.]{dosovitskiy2020image}
Alexey Dosovitskiy, Lucas Beyer, Alexander Kolesnikov, Dirk Weissenborn,
  Xiaohua Zhai, Thomas Unterthiner, Mostafa Dehghani, Matthias Minderer, Georg
  Heigold, Sylvain Gelly, et~al.
\newblock An image is worth 16x16 words: Transformers for image recognition at
  scale.
\newblock \emph{arXiv preprint arXiv:2010.11929}, 2020.

\bibitem[Fan et~al.(2019)Fan, Chen, Kuehne, Pistoia, and Cox]{fan2019more}
Quanfu Fan, Chun-Fu Chen, Hilde Kuehne, Marco Pistoia, and David Cox.
\newblock More is less: Learning efficient video representations by big-little
  network and depthwise temporal aggregation.
\newblock \emph{arXiv preprint arXiv:1912.00869}, 2019.

\bibitem[Feichtenhofer(2020)]{feichtenhofer2020x3d}
Christoph Feichtenhofer.
\newblock X3d: Expanding architectures for efficient video recognition.
\newblock In \emph{Proceedings of the IEEE/CVF Conference on Computer Vision
  and Pattern Recognition}, pages 203--213, 2020.

\bibitem[Feichtenhofer et~al.(2019)Feichtenhofer, Fan, Malik, and
  He]{feichtenhofer2019slowfast}
Christoph Feichtenhofer, Haoqi Fan, Jitendra Malik, and Kaiming He.
\newblock Slowfast networks for video recognition.
\newblock In \emph{Proceedings of the IEEE/CVF International Conference on
  Computer Vision}, pages 6202--6211, 2019.

\bibitem[Goyal et~al.(2017)Goyal, Ebrahimi~Kahou, Michalski, Materzynska,
  Westphal, Kim, Haenel, Fruend, Yianilos, Mueller-Freitag,
  et~al.]{goyal2017something}
Raghav Goyal, Samira Ebrahimi~Kahou, Vincent Michalski, Joanna Materzynska,
  Susanne Westphal, Heuna Kim, Valentin Haenel, Ingo Fruend, Peter Yianilos,
  Moritz Mueller-Freitag, et~al.
\newblock The" something something" video database for learning and evaluating
  visual common sense.
\newblock In \emph{Proceedings of the IEEE International Conference on Computer
  Vision}, pages 5842--5850, 2017.

\bibitem[He et~al.(2016)He, Zhang, Ren, and Sun]{he2016deep}
Kaiming He, Xiangyu Zhang, Shaoqing Ren, and Jian Sun.
\newblock Deep residual learning for image recognition.
\newblock In \emph{Proceedings of the IEEE conference on computer vision and
  pattern recognition}, pages 770--778, 2016.

\bibitem[Jiang et~al.(2019)Jiang, Wang, Gan, Wu, and Yan]{jiang2019stm}
Boyuan Jiang, MengMeng Wang, Weihao Gan, Wei Wu, and Junjie Yan.
\newblock Stm: Spatiotemporal and motion encoding for action recognition.
\newblock In \emph{Proceedings of the IEEE/CVF International Conference on
  Computer Vision}, pages 2000--2009, 2019.

\bibitem[Karpathy et~al.(2014)Karpathy, Toderici, Shetty, Leung, Sukthankar,
  and Fei-Fei]{karpathy2014large}
Andrej Karpathy, George Toderici, Sanketh Shetty, Thomas Leung, Rahul
  Sukthankar, and Li~Fei-Fei.
\newblock Large-scale video classification with convolutional neural networks.
\newblock In \emph{Proceedings of the IEEE conference on Computer Vision and
  Pattern Recognition}, pages 1725--1732, 2014.

\bibitem[Kay et~al.(2017)Kay, Carreira, Simonyan, Zhang, Hillier,
  Vijayanarasimhan, Viola, Green, Back, Natsev, et~al.]{kay2017kinetics}
Will Kay, Joao Carreira, Karen Simonyan, Brian Zhang, Chloe Hillier, Sudheendra
  Vijayanarasimhan, Fabio Viola, Tim Green, Trevor Back, Paul Natsev, et~al.
\newblock The kinetics human action video dataset.
\newblock \emph{arXiv preprint arXiv:1705.06950}, 2017.

\bibitem[Kazakos et~al.(2019)Kazakos, Nagrani, Zisserman, and
  Damen]{kazakos2019epic}
Evangelos Kazakos, Arsha Nagrani, Andrew Zisserman, and Dima Damen.
\newblock Epic-fusion: Audio-visual temporal binding for egocentric action
  recognition.
\newblock In \emph{Proceedings of the IEEE/CVF International Conference on
  Computer Vision}, pages 5492--5501, 2019.

\bibitem[Krizhevsky et~al.(2012)Krizhevsky, Sutskever, and
  Hinton]{krizhevsky2012imagenet}
Alex Krizhevsky, Ilya Sutskever, and Geoffrey~E Hinton.
\newblock Imagenet classification with deep convolutional neural networks.
\newblock \emph{Advances in neural information processing systems},
  25:\penalty0 1097--1105, 2012.

\bibitem[Kwon et~al.(2020)Kwon, Kim, Kwak, and Cho]{kwon2020motionsqueeze}
Heeseung Kwon, Manjin Kim, Suha Kwak, and Minsu Cho.
\newblock Motionsqueeze: Neural motion feature learning for video
  understanding.
\newblock In \emph{European Conference on Computer Vision}, pages 345--362.
  Springer, 2020.

\bibitem[Li et~al.(2020)Li, Ji, Shi, Zhang, Kang, and Wang]{li2020tea}
Yan Li, Bin Ji, Xintian Shi, Jianguo Zhang, Bin Kang, and Limin Wang.
\newblock Tea: Temporal excitation and aggregation for action recognition.
\newblock In \emph{Proceedings of the IEEE/CVF Conference on Computer Vision
  and Pattern Recognition}, pages 909--918, 2020.

\bibitem[Lin et~al.(2019)Lin, Gan, and Han]{lin2019tsm}
Ji~Lin, Chuang Gan, and Song Han.
\newblock Tsm: Temporal shift module for efficient video understanding.
\newblock In \emph{Proceedings of the IEEE/CVF International Conference on
  Computer Vision}, pages 7083--7093, 2019.

\bibitem[Liu et~al.(2020)Liu, Wang, Wu, Qian, and Lu]{liu2020tam}
Zhaoyang Liu, Limin Wang, Wayne Wu, Chen Qian, and Tong Lu.
\newblock Tam: Temporal adaptive module for video recognition.
\newblock \emph{arXiv preprint arXiv:2005.06803}, 2020.

\bibitem[Paszke et~al.(2019)Paszke, Gross, Massa, Lerer, Bradbury, Chanan,
  Killeen, Lin, Gimelshein, Antiga, et~al.]{paszke2019pytorch}
Adam Paszke, Sam Gross, Francisco Massa, Adam Lerer, James Bradbury, Gregory
  Chanan, Trevor Killeen, Zeming Lin, Natalia Gimelshein, Luca Antiga, et~al.
\newblock Pytorch: An imperative style, high-performance deep learning library.
\newblock \emph{arXiv preprint arXiv:1912.01703}, 2019.

\bibitem[Qiu et~al.(2019)Qiu, Yao, Ngo, Tian, and Mei]{qiu2019learning}
Zhaofan Qiu, Ting Yao, Chong-Wah Ngo, Xinmei Tian, and Tao Mei.
\newblock Learning spatio-temporal representation with local and global
  diffusion.
\newblock In \emph{Proceedings of the IEEE/CVF Conference on Computer Vision
  and Pattern Recognition}, pages 12056--12065, 2019.

\bibitem[Raffel et~al.(2019)Raffel, Shazeer, Roberts, Lee, Narang, Matena,
  Zhou, Li, and Liu]{raffel2019exploring}
Colin Raffel, Noam Shazeer, Adam Roberts, Katherine Lee, Sharan Narang, Michael
  Matena, Yanqi Zhou, Wei Li, and Peter~J Liu.
\newblock Exploring the limits of transfer learning with a unified text-to-text
  transformer.
\newblock \emph{arXiv preprint arXiv:1910.10683}, 2019.

\bibitem[Srinivas et~al.(2021)Srinivas, Lin, Parmar, Shlens, Abbeel, and
  Vaswani]{srinivas2021bottleneck}
Aravind Srinivas, Tsung-Yi Lin, Niki Parmar, Jonathon Shlens, Pieter Abbeel,
  and Ashish Vaswani.
\newblock Bottleneck transformers for visual recognition.
\newblock \emph{arXiv preprint arXiv:2101.11605}, 2021.

\bibitem[Touvron et~al.(2020)Touvron, Cord, Douze, Massa, Sablayrolles, and
  J{\'e}gou]{touvron2020training}
Hugo Touvron, Matthieu Cord, Matthijs Douze, Francisco Massa, Alexandre
  Sablayrolles, and Herv{\'e} J{\'e}gou.
\newblock Training data-efficient image transformers \& distillation through
  attention.
\newblock \emph{arXiv preprint arXiv:2012.12877}, 2020.

\bibitem[Tran et~al.(2015)Tran, Bourdev, Fergus, Torresani, and
  Paluri]{tran2015learning}
Du~Tran, Lubomir Bourdev, Rob Fergus, Lorenzo Torresani, and Manohar Paluri.
\newblock Learning spatiotemporal features with 3d convolutional networks.
\newblock In \emph{Proceedings of the IEEE international conference on computer
  vision}, pages 4489--4497, 2015.

\bibitem[Tran et~al.(2018)Tran, Wang, Torresani, Ray, LeCun, and
  Paluri]{tran2018closer}
Du~Tran, Heng Wang, Lorenzo Torresani, Jamie Ray, Yann LeCun, and Manohar
  Paluri.
\newblock A closer look at spatiotemporal convolutions for action recognition.
\newblock In \emph{Proceedings of the IEEE conference on Computer Vision and
  Pattern Recognition}, pages 6450--6459, 2018.

\bibitem[Tran et~al.(2019)Tran, Wang, Torresani, and Feiszli]{tran2019video}
Du~Tran, Heng Wang, Lorenzo Torresani, and Matt Feiszli.
\newblock Video classification with channel-separated convolutional networks.
\newblock In \emph{Proceedings of the IEEE/CVF International Conference on
  Computer Vision}, pages 5552--5561, 2019.

\bibitem[Vaswani et~al.(2017)Vaswani, Shazeer, Parmar, Uszkoreit, Jones, Gomez,
  Kaiser, and Polosukhin]{vaswani2017attention}
Ashish Vaswani, Noam Shazeer, Niki Parmar, Jakob Uszkoreit, Llion Jones,
  Aidan~N Gomez, Lukasz Kaiser, and Illia Polosukhin.
\newblock Attention is all you need.
\newblock \emph{arXiv preprint arXiv:1706.03762}, 2017.

\bibitem[Wang et~al.(2020{\natexlab{a}})Wang, Tran, Torresani, and
  Feiszli]{wang2020video}
Heng Wang, Du~Tran, Lorenzo Torresani, and Matt Feiszli.
\newblock Video modeling with correlation networks.
\newblock In \emph{Proceedings of the IEEE/CVF Conference on Computer Vision
  and Pattern Recognition}, pages 352--361, 2020{\natexlab{a}}.

\bibitem[Wang et~al.(2016)Wang, Xiong, Wang, Qiao, Lin, Tang, and
  Van~Gool]{wang2016temporal}
Limin Wang, Yuanjun Xiong, Zhe Wang, Yu~Qiao, Dahua Lin, Xiaoou Tang, and Luc
  Van~Gool.
\newblock Temporal segment networks: Towards good practices for deep action
  recognition.
\newblock In \emph{European conference on computer vision}, pages 20--36.
  Springer, 2016.

\bibitem[Wang et~al.(2018{\natexlab{a}})Wang, Xiong, Wang, Qiao, Lin, Tang, and
  Van~Gool]{wang2018temporal}
Limin Wang, Yuanjun Xiong, Zhe Wang, Yu~Qiao, Dahua Lin, Xiaoou Tang, and Luc
  Van~Gool.
\newblock Temporal segment networks for action recognition in videos.
\newblock \emph{IEEE transactions on pattern analysis and machine
  intelligence}, 41\penalty0 (11):\penalty0 2740--2755, 2018{\natexlab{a}}.

\bibitem[Wang et~al.(2020{\natexlab{b}})Wang, Xiong, Neumann, Piergiovanni,
  Ryoo, Angelova, Kitani, and Hua]{wang2020attentionnas}
Xiaofang Wang, Xuehan Xiong, Maxim Neumann, AJ~Piergiovanni, Michael~S Ryoo,
  Anelia Angelova, Kris~M Kitani, and Wei Hua.
\newblock Attentionnas: Spatiotemporal attention cell search for video
  classification.
\newblock In \emph{European Conference on Computer Vision}, pages 449--465.
  Springer, 2020{\natexlab{b}}.

\bibitem[Wang et~al.(2018{\natexlab{b}})Wang, Girshick, Gupta, and
  He]{wang2018non}
Xiaolong Wang, Ross Girshick, Abhinav Gupta, and Kaiming He.
\newblock Non-local neural networks.
\newblock In \emph{Proceedings of the IEEE conference on computer vision and
  pattern recognition}, pages 7794--7803, 2018{\natexlab{b}}.

\bibitem[Wu et~al.(2018)Wu, Wan, Yue, Jin, Zhao, Golmant, Gholaminejad,
  Gonzalez, and Keutzer]{wu2018shift}
Bichen Wu, Alvin Wan, Xiangyu Yue, Peter Jin, Sicheng Zhao, Noah Golmant, Amir
  Gholaminejad, Joseph Gonzalez, and Kurt Keutzer.
\newblock Shift: A zero flop, zero parameter alternative to spatial
  convolutions.
\newblock In \emph{CVPR}, 2018.

\bibitem[Wu et~al.(2020)Wu, Girshick, He, Feichtenhofer, and
  Krahenbuhl]{wu2020multigrid}
Chao-Yuan Wu, Ross Girshick, Kaiming He, Christoph Feichtenhofer, and Philipp
  Krahenbuhl.
\newblock A multigrid method for efficiently training video models.
\newblock In \emph{Proceedings of the IEEE/CVF Conference on Computer Vision
  and Pattern Recognition}, pages 153--162, 2020.

\bibitem[Xie et~al.(2017)Xie, Girshick, Doll{\'a}r, Tu, and
  He]{xie2017aggregated}
Saining Xie, Ross Girshick, Piotr Doll{\'a}r, Zhuowen Tu, and Kaiming He.
\newblock Aggregated residual transformations for deep neural networks.
\newblock In \emph{Proceedings of the IEEE conference on computer vision and
  pattern recognition}, pages 1492--1500, 2017.

\bibitem[Yuan et~al.(2021)Yuan, Chen, Wang, Yu, Shi, Tay, Feng, and
  Yan]{yuan2021tokens}
Li~Yuan, Yunpeng Chen, Tao Wang, Weihao Yu, Yujun Shi, Francis~EH Tay, Jiashi
  Feng, and Shuicheng Yan.
\newblock Tokens-to-token vit: Training vision transformers from scratch on
  imagenet.
\newblock \emph{arXiv preprint arXiv:2101.11986}, 2021.

\bibitem[Zhang et~al.(2019)Zhang, Goodfellow, Metaxas, and
  Odena]{zhang2019self}
Han Zhang, Ian Goodfellow, Dimitris Metaxas, and Augustus Odena.
\newblock Self-attention generative adversarial networks.
\newblock In \emph{International conference on machine learning}, pages
  7354--7363. PMLR, 2019.

\bibitem[Zhou et~al.(2018)Zhou, Andonian, Oliva, and
  Torralba]{zhou2018temporal}
Bolei Zhou, Alex Andonian, Aude Oliva, and Antonio Torralba.
\newblock Temporal relational reasoning in videos.
\newblock In \emph{Proceedings of the European Conference on Computer Vision
  (ECCV)}, pages 803--818, 2018.

\end{thebibliography}
}

\end{document}